\documentclass[
]{ceurart}

\usepackage{listings}
\usepackage[utf8]{inputenc}
\usepackage[T1]{fontenc}
\usepackage{amsmath}
\usepackage{amssymb}
\usepackage{graphicx}
\usepackage{tikz}
\usetikzlibrary{matrix}
\usetikzlibrary{positioning}
\usepackage{amsfonts}
\usepackage{hyperref}
\usepackage{verbatim}
\newtheorem{definition}{Definition}
\usepackage{caption}
\usepackage{subcaption}
\usepackage{mathtools}
\newtheorem{proposition}{Proposition}
\begin{document}

\copyrightyear{2024}
\copyrightclause{Copyright for this paper by its authors.
  Use permitted under Creative Commons License Attribution 4.0
  International (CC BY 4.0).}


\title{Probabilistic Abstract Interpretation on Neural Networks via Grids Approximation}


\author[1]{Zhuofan Zhang}[%
email=zhuofan.zhang13@imperial.ac.uk,
]
\address[1]{Imperial College London,
  180 Queen's Gate, South Kensington, London SW7 2AZ}

\author[2]{Herbert Wiklicky}[%
email=h.wiklicky@imperial.ac.uk,
]
\address[2]{Imperial College London,
  180 Queen's Gate, South Kensington, London SW7 2AZ}


\begin{abstract}
Probabilistic abstract interpretation is a theory used to extract particular properties of a computer program when it is infeasible to test every single inputs. In this paper we apply the theory on neural networks for the same purpose: to analyse density distribution flow of all possible inputs of a neural network when a network has uncountably many or countable but infinitely many inputs. We show how this theoretical framework works in neural networks and then discuss different abstract domains and corresponding Moore-Penrose pseudo-inverses together with abstract transformers used in the framework. We also present experimental examples to show how this framework helps to analyse real world problems.
\end{abstract}

\begin{keywords}
  explainable AI \sep
  probabilistic abstract interpretation \sep
  neural network
\end{keywords}


\conference{xaiit2024}

\maketitle

\section{Introduction}

In recent years, deep neural networks has been regarded as an essential technique for machine learning since the work of deep learning is being progressed \cite{deeplearning,dlgoodfellow} and have been playing in a big role in many industries in real world. However, a major problem of it is that it is difficult to interpret how neural networks deal with information and derive their decisions, especially when layers are getting deeper. Interpreting neural networks is important because in some application areas cost of mistakes is intolerable. Our approach aims at making use of the theory of probabilistic abstract interpretation to analyse density distribution flow of possible inputs of a neural network and to make sure the system is safe and under controlled.\\ 
Research on interpreting neural networks has been carried on for years. One approach to do is through rule extraction algorithms \cite{review}. Activation Maximisation is another approach to interpret model prediction by only looking at its relationships with input, disregarding the inner process \cite{am}. A further development of the technique is to add a decoder on it \cite{am_decoder}. In the same year Bazen et al. used Taylor decomposition explain model’s decision by decomposing output as a sum of relevance scores \cite{taylor_decomposition}, and Landecker et al. decomposed the model prediction by back-propagating relevance in reverse direction, progressively redistributing the prediction score until the input is
reached \cite{releprop}. Bach et al. further introduced the Layer-wise Relevance Propagation explanation framework \cite{lrp}. Another measurement is to use sensitivity analysis to quantify the importance of each input variable \cite{samek}. In 2015 Zilke proposed DeepRED, by extending a decompositional algorithm CRED \cite{deepred}. Jacobsson \cite{rnn} and Wang \cite{rnn2} have proposed algorithms of understanding recurrent neural networks. In 2018 Zhang et al. and Zhou et al. respectively proposed methods to interpret convolutional neural networks \cite{Quanshi1, Quanshi2, Bolei}. Zeiler et al. instead used approach of layer activation visualisation with algorithm of Deconvnets to understand CNN \cite{visualize}. Bau et al. proposed Network Dissection framework which well quantifies the interpretability of latent representations of CNN \cite{dissection}. In the mean time, the attacks and defenses from adversarial examples to neural networks are frequently studied \cite{adversarial}.\\
Program analysis, on the other hand, is a research area which focuses on using mathematical theory and formal methods to automatically analyse the behavior of a computer program regarding a property such as safety, correctness, robustness and liveness \cite{nielsonbook}. In particular, abstract interpretation is a theory of approximating semantics of a computer program based on order theory, especially lattices \cite{cousot,AI}. In recent years, probabilistic program analysis has been introduced to analyze quantitative aspect of computer programs. Probabilistic abstract interpretation correspondingly is the framework to lift classical abstract interpretation to the probabilistic setting \cite{paicousot,ccp,psa,PAI}.\\
In 2018 SRILAB proposed $AI^{2}$ analyzer that uses abstract interpretation to over-approximate a range of concrete values into an abstract domain and checks whether this abstract domain holds the expected result, and which will prove every single input over-approximated by this abstract domain also holds this result \cite{ai2}. Later \cite{ai2_2} proposed a new abstract domain which combines floating point polyhedra with intervals. Its performance was compared to Reluplex which is a former property verification technique extending the simplex algorithm to support ReLU constraints \cite{reluplex}. Abstract interpretation is also bridged with gradient-based optimization and can be applied on training neural networks \cite{ai2ontrain}.\\
In this paper, we take theory of probabilistic abstract interpretation to analyse density distribution flow of all possible inputs of a neural network, when the network has uncountably many or countable but infinitely many possible inputs. We also see that in particular situations probabilistic abstract interpretation has stronger features than classical ones. Section 2 gives a brief background of probabilistic abstract interpretation theory. In Section 3 we introduce how to apply the theory on neural networks, and discuss different choices of abstract domains for different purposes, together with corresponding Moore-Penrose pseudo inverses and abstract domains. In Section 4 we show in experiments how the framework helps in real world problem. Section 5 gives a conclusion and future works.

\section{Background: Probabilistic Abstract Interpretation Theory}

In this section we give a brief introduction of probabilistic abstract interpretation theory originally introduced in program analysis. 
\begin{figure}
\centering
\begin{tikzpicture}[node distance=2cm]
\node(C)                            {$C$};
\node(D)     [below of=C]           {$D$};
\node(Csharp)     [right =4cm of C]     {$C^{\#}$};
\node(Dsharp)     [right =4cm of D]     {$D^{\#}$};

\draw[->](C)         -- node[left]{$f$} (D);
\draw[->](Csharp)         -- node[right]{$f^{\#}$} (Dsharp);
\draw[->](C.15)         -- node[above]{$A$}(Csharp.170);
\draw[<-](C.345)         -- node[below]{$G$}(Csharp.190);
\draw[->](D.15)         -- node[above]{$A'$}(Dsharp.170);
\draw[<-](D.345)         -- node[below]{$G'$}(Dsharp.190);

\end{tikzpicture}
\caption{Probabilistic abstract interpretation with Moore-Penrose pseudo-inverse pairs $A$ and $G$. The left hand side $C \xrightarrow{\text{$f$}} D$ is the concrete domain and the right hand side $C^{\#} \xrightarrow{\text{$f^{\#}$}} D^{\#}$ is the abstract domain.}
\end{figure}
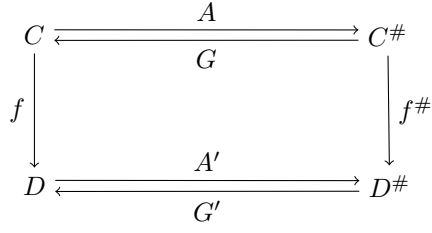
Probabilistic abstract interpretation is considered as a quantitative version of the classical abstract interpretation, commonly to give a not safe but a more average case scenario, while the classical framework keeps safe but might be not precise enough and therefore addressing a worst case scenario \cite{nielsonbook, cousot, ccp, lecturenotes, psa}. In probabilistic framework, instead of forming a Galois connection in which classical framework does, we take a Moore-Penrose pseudo-inverse.
\begin{definition}
Let $C$ and $C^{\#}$ be two finite-dimensional vector (Hilbert) spaces and $A : C \rightarrow C^{\#}$ a linear map. Then the linear map $G : C^{\#} \rightarrow C$ is the Moore-Penrose pseudo-inverse of A iff (i) $A \circ G = P_A$ (ii) $G \circ A = P_G$, where $P_A$ and $P_G$ denote orthogonal projections onto the ranges of $A$ and $G$.
\end{definition}
As shown in Figure 1, $f$ is the original function in concrete domain, which can be a step in a computer program. Based on $f$, we further consider function $\overrightarrow{f}$ in probability domain and function $f^{\#}$ in abstract domain respectively. Given in concrete domain
\begin{equation}
    f: R \rightarrow R
\end{equation}
we lift $f$ to $\overrightarrow{f}$ which is a mapping from a distribution to a distribution in probabilistic domain
\begin{equation}
    \overrightarrow{f}: \mathcal{V}(R) \rightarrow \mathcal{V}(R)
\end{equation}
where $\mathcal{V}(R)$ is the vector space of $R$, and from probabilistic abstract interpretation framework in Figure 1, we can formalise $f^{\#}$ in abstract domain as
\begin{equation}
    f^{\#} = A' f G
\end{equation}
where we have Moore-Penrose pseudo-inverse as connections between concrete and abstract domains. $f^{\#}$ is also called the \textbf{abstract transformer}. Defining an appropriate abstract domain and formalising an appropriate abstract transformer is crucial to performance of this analysis framework.
\begin{proposition}
An abstract transformer of a composition of functions is a composition of abstract transformers of each function, that is
\begin{equation}
    \text{if } f = f' \circ f'' \text{, then } f^{\#} = f'^{\#} \circ f''^{\#}
\end{equation}
\end{proposition}

\section{Probabilistic Abstract Interpretation on Neural Networks}
In this section we discuss how to apply theory of probabilistic abstract interpretation on neural networks. In principle, any neural network is a non-linear function which can be simulated by a computer program, thus any analysis framework working on a computer program can work on a neural network.
\subsection{Probabilistic Abstract Interpretation on A General Layer}
A typical layer in a feed-forward neural network can be expressed as an affine transformation followed by a non-linear function, or more specifically can be for example in form of 
\begin{equation}
    x^{t+1} = ReLU(W^{t}x^{t} + b^{t})
\end{equation}
where $t$ is the layer number, $x^{t}$ is the input of the layer, $x^{t+1}$ is the output, $W^{t}$ and $b^{t}$ are the weights of affine transformation, and $ReLU$ is one example of choices of activation functions. We use these notations instead of $f$ to avoid ambiguity from $f$ in abstract interpretation. To construct a probabilistic abstract interpretation, $f$, $\overrightarrow{f}$ and $f^{\#}$ need to be considered. In this case $f : \mathbb{R}^m \rightarrow \mathbb{R}^n$ is the forward propagation function. We lift $f : \mathbb{R}^m \rightarrow \mathbb{R}^n$ up into probabilistic domain to be $\overrightarrow{f} : \mathcal{V}(\mathbb{R}^m) \rightarrow \mathcal{V}(\mathbb{R}^n)$ which is
\begin{equation}
    d(x^{t+1}) = \overrightarrow{ReLU}(\overrightarrow{W}^{t}d(x^{t}) + \overrightarrow{b}^{t})
\end{equation}
where $d(x)$ is probability distribution of $x$. Then from $\overrightarrow{f}$ we can construct the abstract transformer $f^{\#}$ which is
\begin{equation}
    d^{\#}(x^{t+1}) = ReLU^{\#}(W^{t\#}d^{\#}(x^{t}) + b^{t\#})
\end{equation}
In neural networks it is very common for a layer to be a composition of multiple functions. A typical case will be an affine transformation followed by a non-linear activation function. To obtain the abstract transformer, by definition of Moore-Penrose pseudo-inverse we can have abstract transformer for Aff$^t$
\begin{equation}
    \text{Aff}^{t\#} = A' \overrightarrow{\text{Aff}^{t}} G
\end{equation}
and abstract transformer for $ReLU$
\begin{equation}
    ReLU^{\#} = A'' \overrightarrow{ReLU} G'
\end{equation}
and by Proposition 1 we can have
\begin{equation}
    (\text{Aff}^t \circ ReLU)^{\#} = \text{Aff}^{t\#} \circ ReLU^{\#} 
\end{equation}
In probabilistic abstract interpretation, each possible value a neuron has is with a corresponding probability. A prior distribution initialisation is needed to be given in input space. We use tensor product to combine all probability distributions for every single neurons in input space to form an overall distribution representing the input layer. That is, for any two independent neurons,
\begin{equation}
    \mathcal{V}(X^{1}_{1} \times X^{1}_{2}) = \mathcal{V}(X^{1}_{1}) \otimes \mathcal{V}(X^{1}_{2})
\end{equation}
where $\otimes$ is notation of tensor product. Therefore, the abstract transformer for the first layer $f^{1\#}$ is
\begin{equation}
         (\text{Aff}^t \circ ReLU)^{\#}(x^1) 
         = ReLU^{\#}(W^{1\#}(d(x^1_1) \otimes ... \otimes d(x^1_n)) + b^{1\#})  
\end{equation}
\subsection{Probabilistic Abstract Interpretation on Basic MLP}
Given a simplest MLP, with two layers and five neurons, with parameters shown below. We use notations $x^{t}$ to represent neuron values in layer $t$. Let inputs to be real values such that $x_1^1, x_2^1 \in [-3, 3]$, 
\begin{equation}
W^{1} = 
    \begin{pmatrix}
    1 & 1 \\
    1 & 1 
    \end{pmatrix}, 
b^{1} = 
    \begin{pmatrix}
    0 \\
    0
    \end{pmatrix}, 
W^{2} = 
    \begin{pmatrix}
    1 \\   
    1
    \end{pmatrix}, 
b^{2} = 0
\end{equation}
and we set both activation functions as ReLU.
\begin{figure}
\centering
\begin{tikzpicture}[scale=0.6]
\title{lattice of abstract domain}
\node(top)                           {$\top$};
\node(zero)     [below of=top]       {$0$};
\node(pos)      [left  of=zero]      {$\mathbb{R}^{+}$};
\node(neg)      [right of=zero]      {$\mathbb{R}^{-}$};
\node(bot)      [below of=zero]      {$\bot$};

\draw(top)       -- (pos);
\draw(top)       -- (zero);
\draw(top)       -- (neg);
\draw(pos)       -- (bot);
\draw(zero)      -- (bot);
\draw(neg)       -- (bot);

\end{tikzpicture}
\caption{lattice of abstract domain}
\end{figure}
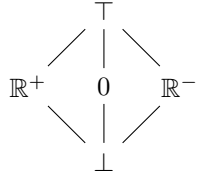
We take the abstract domain to be $\{\mathbb{R}^{+}, 0, \mathbb{R}^{-}\}$, shown in Fig. 2. In concrete domain we have $f : \mathbb{R}^2 \rightarrow \mathbb{R}^2$ 
\begin{equation}
    (x^2_1, x^2_2) = ReLU(W^1(x^1_1, x^1_2) + b^1)
\end{equation}
and we have $\overrightarrow{f} : \mathcal{V}(\mathbb{R}^2) \rightarrow \mathcal{V}(\mathbb{R}^2)$
\begin{equation}
    d(x^2) = \overrightarrow{ReLU}(\overrightarrow{\text{Aff}^1}(d(x^1_1) \otimes d(x^1_2))) 
\end{equation}
In abstract domain, suppose we first only discuss the inputs in integers $x_1^1, x_2^1 \in \mathbb{N}$ such that 
\begin{equation}
x^1_1, x^1_2 \in [-3, -2, -1, 0, 1, 2, 3]
\end{equation}
and initialise input space with uniform distribution, we have 
\begin{equation}
    d(x^1_1) = d(x^1_2) = (3/7, 1/7, 3/7) 
\end{equation}
and we can have 
\begin{equation}
    \begin{split}
    d(x^2) &= ReLU^{\#}(\text{Aff}^{1\#}((3/7, 1/7, 3/7)  \otimes (3/7, 1/7, 3/7) )))\\
    &= \begin{pmatrix}
0 & 0 & 0 & 0 & 4/7 & 0 & 0 & 0 & 3/7\\
\end{pmatrix}
    \end{split}
\end{equation}
That is, after first layer, we will have $(x_1^2 = 0, x_2^2 = 0)$ with probability 4/7, and $(x_1^2 \in \mathbb{R}^{+}, x_2^2 \in \mathbb{R}^{+})$ with probability 3/7, and zero probability to have other outcomes. To be more specific, we have 
\begin{equation}
    \overrightarrow{\text{Aff}^1} = \begin{pmatrix}
1 & 0 & 0 & 0 & 0 & 0 & 0 & 0 & 0 & \cdots & 0\\ 
0 & 1 & 0 & 0 & 0 & 0 & 0 & 1 & 0 & \cdots & 0 \\ 
\vdots&\vdots&\vdots&\vdots&\vdots&\vdots&\vdots&\vdots&\vdots&\ddots&\vdots\\
0 & 0 & 0 & 0 & 0 & 0 & 0 & 0 & 0 & \cdots & 1 \\
\end{pmatrix}
\in \mathbb{R}^{13 \times 49}
\end{equation}
where in a row each entry represents a possible pair of $(x_1^1, x_2^1)$, and in a column each entry represents entry of $x_1^1 + x_2^1$. For example, in the 1st row of $\overrightarrow{\text{Aff}^1}$, only first entry is 1 and all others are zero. That is, there exists and only exists a possible pair $(-3, -3)$ that will result in $x_1^1 + x_2^1 = -6$ which the 1st row represents. We also have 
\begin{equation}
    A' = \begin{pmatrix}
1 & 1 & 1 & 1 & 1 & 1 & 0 & 0 & 0 & 0 & 0 & 0 & 0\\ 
0 & 0 & 0 & 0 & 0 & 0 & 0 & 0 & 0 & 0 & 0 & 0 & 0\\
\vdots&\vdots&\vdots&\vdots&\vdots&\vdots&\vdots&\vdots&\vdots&\vdots&\vdots&\vdots&\vdots\\
0 & 0 & 0 & 0 & 0 & 0 & 1 & 0 & 0 & 0 & 0 & 0 & 0\\
\vdots&\vdots&\vdots&\vdots&\vdots&\vdots&\vdots&\vdots&\vdots&\vdots&\vdots&\vdots&\vdots\\
0 & 0 & 0 & 0 & 0 & 0 & 0 & 0 & 0 & 0 & 0 & 0 & 0\\
0 & 0 & 0 & 0 & 0 & 0 & 0 & 1 & 1 & 1 & 1 & 1 & 1\\
\end{pmatrix}
\in \mathbb{R}^{9 \times 13}
\end{equation}
which in output space maps a probability distribution vector to an abstract element vector, and 
\begin{equation}
    G = \begin{pmatrix}
\frac{1}{9} & 0 & 0 & 0 & 0 & 0 & 0 & 0 & 0\\ 
\frac{1}{9} & 0 & 0 & 0 & 0 & 0 & 0 & 0 & 0\\ 
\vdots&\vdots&\vdots&\vdots&\vdots&\vdots&\vdots&\vdots&\vdots\\
0 & 0 & 0 & 0 & 0 & 0 & 0 & 0 & \frac{1}{9}\\ 
\end{pmatrix}
\in \mathbb{R}^{49 \times 9} 
\end{equation}
which in input space maps an abstract element vector to a probability distribution vector. As part of information is lost, we can only map back by uniform distribution. For example, there are nine possible pairs of $(x_1^1, x_2^1)$ with both values negative, so we let the total probability of $(-, -)$ divided by nine to be probability of each possible pair. Then, recall definition of probabilistic abstract interpretation, we have 
\begin{equation}
    \text{Aff}^{1\#} = A'\overrightarrow{\text{Aff}^1}G
\end{equation}
and similarly we can have
\begin{equation}
    ReLU^{\#} = A''\overrightarrow{ReLU}G'
\end{equation}
which switches probabilities in positions of negativeness into positions of zeros. In this case, it is easy to obtain $ReLU^{\#}$
\begin{equation}
    ReLU^{\#} = \begin{pmatrix}
0 & 0 & 0 & 0 & 0 & 0 & 0 & 0 & 0\\ 
0 & 0 & 0 & 0 & 0 & 0 & 0 & 0 & 0\\ 
0 & 0 & 0 & 0 & 0 & 0 & 0 & 0 & 0\\ 
0 & 0 & 0 & 0 & 0 & 0 & 0 & 0 & 0\\ 
1 & 1 & 0 & 1 & 1 & 0 & 0 & 0 & 0\\ 
0 & 0 & 1 & 0 & 0 & 1 & 0 & 0 & 0\\ 
0 & 0 & 0 & 0 & 0 & 0 & 0 & 0 & 0\\
0 & 0 & 0 & 0 & 0 & 0 & 1 & 1 & 0\\
0 & 0 & 0 & 0 & 0 & 0 & 0 & 0 & 1\\ 
\end{pmatrix}
\in \mathbb{R}^{9 \times 9} 
\end{equation}
\\Then to make the case more general if we expand the input domain into reals $x_1^1, x_2^1 \in \mathbb{R}$ such that 
\begin{equation}
x^1_1, x^1_2 \in [-3, 3]
\end{equation}
we can have 
\begin{equation}
    d(x^1_1) = d(x^1_2) = (1/2, 0, 1/2) 
\end{equation}
and 
\begin{equation}
    \overrightarrow{\text{Aff}^1} \in \lim_{M \rightarrow \infty} \lim_{N \rightarrow \infty} \mathbb{R}^{N \times M}
\end{equation}
which switches positions of pairs after linear transformation, and as in Real space $M$ and $N$ are both very large numbers that tend to infinity, and $M > N$. Also we have 
\begin{equation}
    A' \in \lim_{N \rightarrow \infty} \mathbb{R}^{9 \times N}
\end{equation}
which classifies concrete points in Real space into nine categories of abstract domain, and 
\begin{equation}
    G \in \lim_{M \rightarrow \infty} \mathbb{R}^{M \times 9}
\end{equation}
which maps nine categories of abstract domain back to concrete domain. Hence we have 
\begin{equation}
    \text{Aff}^{1\#} = A'\overrightarrow{\text{Aff}^1}G 
\end{equation}
and 
\begin{equation}
    ReLU^{\#} = A''\overrightarrow{ReLU}G'
\end{equation}
Still, in this particular case
\begin{equation}
    ReLU^{\#} = \begin{pmatrix}
0 & 0 & 0 & 0 & 0 & 0 & 0 & 0 & 0\\ 
0 & 0 & 0 & 0 & 0 & 0 & 0 & 0 & 0\\ 
0 & 0 & 0 & 0 & 0 & 0 & 0 & 0 & 0\\ 
0 & 0 & 0 & 0 & 0 & 0 & 0 & 0 & 0\\ 
1 & 1 & 0 & 1 & 1 & 0 & 0 & 0 & 0\\ 
0 & 0 & 1 & 0 & 0 & 1 & 0 & 0 & 0\\ 
0 & 0 & 0 & 0 & 0 & 0 & 0 & 0 & 0\\
0 & 0 & 0 & 0 & 0 & 0 & 1 & 1 & 0\\
0 & 0 & 0 & 0 & 0 & 0 & 0 & 0 & 1\\ 
\end{pmatrix}
\in \mathbb{R}^{9 \times 9} 
\end{equation}
Therefore
\begin{equation}
    \begin{split}
    d(x^2) &= ReLU^{\#}(\text{Aff}^{1\#}((1/2, 0, 1/2)  \otimes (1/2, 0, 1/2) ))) \\
    &= \begin{pmatrix}
0 & 0 & 0 & 0 & 1/2 & 0 & 0 & 0 & 1/2\\
\end{pmatrix}
    \end{split}
\end{equation}
That is, after first layer, we will have $(x_1^2 = 0, x_2^2 = 0)$ with probability 1/2, and $(x_1^2 \in \mathbb{R}^{+}, x_2^2 \in \mathbb{R}^{+})$ with probability 1/2, and probability 0 to have other outcomes.

\subsection{Probabilistic Abstract Interpretation on Convolutional Layer}
Convolutional neural network in principle is also a special case of fully-connected layers, except that many weights are zeros. We illustrate a simple example of applying abstract interpretation on convolutional layer. Let the input space $X \in \mathbb{R}^{4 \times 4}$, the convolution filter $g \in \mathbb{R}^{2 \times 2}$, and the output space then is $Y \in \mathbb{R}^{(4-2+1) \times (4-2+1)} = \mathbb{R}^{3 \times 3}$. As a well-trained model, the weights of $g$ is already given and fixed. Let
\begin{equation}
    g = 
        \begin{pmatrix}
        g_{11} & g_{12} \\
        g_{21} & g_{22} 
    \end{pmatrix}
\end{equation}
To express the convolution computation in formal mathematical expression, we can have
\begin{equation}
    y = ReLU(Wx + b)
\end{equation}
where $W = \begin{psmallmatrix}
    g_{11} & g_{12} & 0 & 0 & g_{21} & g_{22} & 0 & 0 & 0 & 0 & 0 & 0 & 0 & 0 & 0 & 0 \\
    0 & g_{11} & g_{12} & 0 & 0 & g_{21} & g_{22} & 0 & 0 & 0 & 0 & 0 & 0 & 0 & 0 & 0 \\
    0 & 0 & g_{11} & g_{12} & 0 & 0 & g_{21} & g_{22} & 0 & 0 & 0 & 0 & 0 & 0 & 0 & 0 \\
    0 & 0 & 0 & 0 & g_{11} & g_{12} & 0 & 0 & g_{21} & g_{22} & 0 & 0 & 0 & 0 & 0 & 0  \\
    0 & 0 & 0 & 0 & 0 & g_{11} & g_{12} & 0 & 0 & g_{21} & g_{22} & 0 & 0 & 0 & 0 & 0  \\
    0 & 0 & 0 & 0 & 0 & 0 & g_{11} & g_{12} & 0 & 0 & g_{21} & g_{22} & 0 & 0 & 0 & 0 \\
    0 & 0 & 0 & 0 & 0 & 0 & 0 & 0 & g_{11} & g_{12} & 0 & 0 & g_{21} & g_{22} & 0 & 0  \\
    0 & 0 & 0 & 0 & 0 & 0 & 0 & 0 & 0 & g_{11} & g_{12} & 0 & 0 & g_{21} & g_{22} & 0 \\
    0 & 0 & 0 & 0 & 0 & 0 & 0 & 0 & 0 & 0 & g_{11} & g_{12} & 0 & 0 & g_{21} & g_{22}
\end{psmallmatrix}$

and we choose ReLU function to be activation function as it is the most commonly used one for Convolutional layers. For simplicty we let $b$ be zero vector, and let 
\begin{equation}
    g = 
    \begin{pmatrix}
        1 & -1 \\
        1 & -1
    \end{pmatrix}
\end{equation}
We again concern the positiveness/negativeness, which in computer vision application is the darkness of pixels. The choice of convolution filter is typically a simple version of border detection filter. Following the ReLU function, we can let abstract domain be $\{\mathbb{R}^{-}, 0, \mathbb{R}^{+}\}$. In concrete domain we have $f : \mathbb{R}^{16} \rightarrow \mathbb{R}^{9}$
\begin{equation}
    (y_1, ..., y_9) = ReLU(W(x_1, ..., x_{16}) + b)
\end{equation}
suppose we initialise the input space $X$ with uniform distribution over $\mathbb{R}$, that is, $d(x_i) = (1/2, 0, 1/2)$. To lift up to probability domain we have $\overrightarrow{f} : \mathcal{V}(\mathbb{R}^{16}) \rightarrow \mathcal{V}(\mathbb{R}^{9})$
\begin{equation}
    d(y) = \overrightarrow{ReLU}(\overrightarrow{\text{Aff}}(d(x_1)\otimes ... \otimes d(x_{16})))
\end{equation}
Concretely we can see after convolution each entry has probability 1/2 to be negative and 1/2 to be positive. Therefore, applying ReLU function 
\begin{equation}
    d(y_i) = (0, 1/2, 1/2)
\end{equation}
the output will have probability 1/2 to be 0 and probability 1/2 to be positive consequently, as ReLU will transform all negative values into zero.

\subsection{Zonotopes Approximation as Abstract Domain}
In 2018, Gehr et al introduced $AI^2$, an analyzer which can automatically prove properties, especially safety and robustness, for convolutional neural networks based on classical abstract interpretation \cite{ai2}. $AI^2$ has found a general way to define abstract domain and abstract transformer of different types of layers, including fully-connected layer, Convolutional layer and max-pooling layer. It defines suitable abstract domains and represents neural network layers as CAT functions, and hence defines abstract transformers which can be applied and computed on abstract elements in a very general way. The formal definitions and details can be found in \cite{ai2}.\\
It uses a shape of Zonotope as an abstract domain. In probabilistic framework, we can also follow this abstract domain. We take the following example in \cite{ai2} to illustrate. Given function $f: \mathbb{R}^2 \rightarrow \mathbb{R}^2$
\begin{equation}
    f(x) = 
    \begin{pmatrix}
        2 & -1\\
        0 & 1
    \end{pmatrix}
    x
\end{equation}
with input space X expressed by zonotope $z^1:[-1, 1]^3 \rightarrow \mathbb{R}^2:$
\begin{equation}
    z^1(\epsilon_1, \epsilon_2, \epsilon_3) = (1 + 0.5\epsilon_1 + 0.5\epsilon_2, 2 + 0.5\epsilon_1 + 0.5\epsilon_3)
\end{equation}
followed by a ReLU transformation, shown in Figure 3 and 4. We illustrate this zonotope approximation first in integer space and then generalise it into real space.
\begin{figure}[h]
\centering
\begin{tikzpicture}[scale=0.5]
\draw[fill=gray!22] (0,1) -- (1,1) -- (2,2) -- (2,3) -- (1,3) -- (0,2) -- (0,1);
\draw[fill=gray!22] (11,1) -- (12,2) -- (11,3) -- (9,3) -- (8,2) -- (9,1) -- (11,1);

\draw[->] (-2,0) -- (3,0) coordinate (x axis);
\draw[->] (0,-1) -- (0,4) coordinate (y axis);
\draw[->] (7,0) -- (13,0) coordinate (x axis);
\draw[->] (10,-1) -- (10,4) coordinate (y axis);

\draw[->] (4,2) -- (6,2);

\draw (0,1) -- (0,1) node[anchor=east] {$1$};
\draw (0,2) -- (0,2) node[anchor=east] {$2$};
\draw (0,3) -- (0,3) node[anchor=east] {$3$};
\draw (10,1) -- (10,1) node[anchor=east] {$1$};
\draw (10,2) -- (10,2) node[anchor=east] {$2$};
\draw (10,3) -- (10,3) node[anchor=east] {$3$};

\draw (-1,0) -- (-1,0) node[anchor=north] {$-1$};
\draw (1,0) -- (1,0) node[anchor=north] {$1$};
\draw (2,0) -- (2,0) node[anchor=north] {$2$};
\draw (8,0) -- (8,0) node[anchor=north] {$-2$};
\draw (9,0) -- (9,0) node[anchor=north] {$-1$};
\draw (11,0) -- (11,0) node[anchor=north] {$1$};
\draw (12,0) -- (12,0) node[anchor=north] {$2$};

\end{tikzpicture}
\caption{abstract elements through layer}
\end{figure}
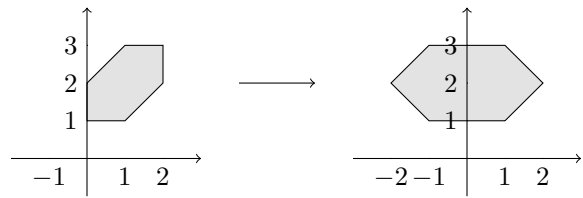
\begin{figure}[h]
\centering
\begin{tikzpicture}[scale=0.5]
\draw[fill=gray!22] (1,1) -- (2,2) -- (1,3) -- (-1,3) -- (-2,2) -- (-1,1) -- (1,1);
\draw[fill=gray!22] (11,1) -- (12,2) -- (11,3) -- (10,3) -- (10,1) -- (11,1);
\draw[-] (10,3) -- (9,2);
\draw[-] (9,2) -- (10,1);

\draw[->] (-3,0) -- (3,0) coordinate (x axis);
\draw[->] (0,-1) -- (0,4) coordinate (y axis);
\draw[->] (7,0) -- (13,0) coordinate (x axis);
\draw[->] (10,-1) -- (10,4) coordinate (y axis);

\draw[->] (4,2) -- (6,2);

\draw (0,1) -- (0,1) node[anchor=east] {$1$};
\draw (0,2) -- (0,2) node[anchor=east] {$2$};
\draw (0,3) -- (0,3) node[anchor=east] {$3$};
\draw (10,1) -- (10,1) node[anchor=east] {$1$};
\draw (10,2) -- (10,2) node[anchor=east] {$2$};
\draw (10,3) -- (10,3) node[anchor=east] {$3$};

\draw (-2,0) -- (-2,0) node[anchor=north] {$-2$};
\draw (-1,0) -- (-1,0) node[anchor=north] {$-1$};
\draw (1,0) -- (1,0) node[anchor=north] {$1$};
\draw (2,0) -- (2,0) node[anchor=north] {$2$};
\draw (8,0) -- (8,0) node[anchor=north] {$-2$};
\draw (9,0) -- (9,0) node[anchor=north] {$-1$};
\draw (11,0) -- (11,0) node[anchor=north] {$1$};
\draw (12,0) -- (12,0) node[anchor=north] {$2$};

\draw [line width=0.5mm, blue] (10,1) -- (10,3);

\end{tikzpicture}
\caption{ReLU transformation}
\end{figure}
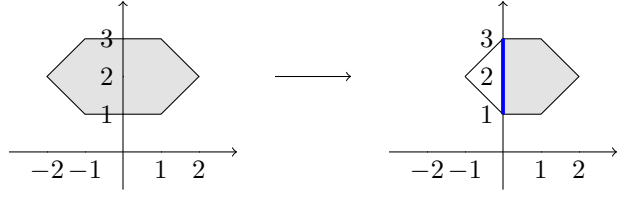

\subsubsection{Zonotopes Approximation in Integer Space}
We first define abstract transformer for probabilistic framework, $f^{\#}$, together with Moore-Penrose pseudo-inverse functions $A$ and $G$ in integer space. In concrete domain, we have overall 4 input points $X = \{(0, 1), (1, 1), (1, 3), (2, 2)\}$ and we lift it up into probability space with uniform distribution to get $d(X) = \{1/4, 1/4, 1/4, 1/4\}$. In abstract domain, the Integer space contained in $z^1$ is $I^2 = \{(0, 1), (0, 2), (1, 1), (1, 2), (1, 3), (2, 2), (2, 3)\}$. As we abstract the input probability space into abstract domain, we have
\begin{equation}
    A(d(X)) = 
    \begin{pmatrix}
        1 & 0 & 0 & 0\\
        0 & 0 & 0 & 0\\
        0 & 1 & 0 & 0\\
        0 & 0 & 0 & 0\\
        0 & 0 & 1 & 0\\
        0 & 0 & 0 & 1\\
        0 & 0 & 0 & 0\\
    \end{pmatrix}
    \begin{pmatrix}
    \frac{1}{4}\\
    \frac{1}{4}\\
    \frac{1}{4}\\
    \frac{1}{4}\\
    \end{pmatrix}
    = 
    \begin{pmatrix}
    \frac{1}{4}\\
    0\\
    \frac{1}{4}\\
    0\\
    \frac{1}{4}\\
    \frac{1}{4}\\
    0\\
    \end{pmatrix}
\end{equation}
and for opposite direction we have
\begin{equation}
    G(A(d(X))) =
    \begin{pmatrix}
        \frac{1}{4} & \frac{1}{4} & \frac{1}{4} & \frac{1}{4} & \frac{1}{4} & \frac{1}{4} & \frac{1}{4}\\
        \frac{1}{4} & \frac{1}{4} & \frac{1}{4} & \frac{1}{4} & \frac{1}{4} & \frac{1}{4} & \frac{1}{4}\\
        \frac{1}{4} & \frac{1}{4} & \frac{1}{4} & \frac{1}{4} & \frac{1}{4} & \frac{1}{4} & \frac{1}{4}\\
        \frac{1}{4} & \frac{1}{4} & \frac{1}{4} & \frac{1}{4} & \frac{1}{4} & \frac{1}{4} & \frac{1}{4}\\
    \end{pmatrix}
    \begin{pmatrix}
    \frac{1}{4}\\
    0\\
    \frac{1}{4}\\
    0\\
    \frac{1}{4}\\
    \frac{1}{4}\\
    0\\
    \end{pmatrix}
    =
    \begin{pmatrix}
    \frac{1}{4}\\
    \frac{1}{4}\\
    \frac{1}{4}\\
    \frac{1}{4}\\
    \end{pmatrix}
\end{equation}
particularly we can also see during linear transformation the distribution over input space does not change, where each point in X corresponds to a point in f(X) and vice versa. Therefore $f^{\#}: \mathcal{V}(\mathbb{R}^2) \rightarrow \mathcal{V}(\mathbb{R}^2)$
\begin{equation}
    f^{\#} = A'\overrightarrow{f}G = A'G
\end{equation}
Similarly the abstract transformer for ReLU is $ReLU^{\#}: \mathcal{V}(\mathbb{R}^2) \rightarrow \mathcal{V}(\mathbb{R}^2)$
\begin{equation}
    \begin{split}
    &ReLU^{\#} = A''\overrightarrow{ReLU}G' = A''G'\\
    &=  \begin{pmatrix}
        0 & 0 & 0 & 0\\
        1 & 0 & 0 & 0\\
        0 & 0 & 0 & 0\\
        0 & 0 & 1 & 0\\
        0 & 1 & 0 & 0\\
        0 & 0 & 0 & 0\\
        0 & 0 & 0 & 0\\
        0 & 0 & 0 & 1\\
    \end{pmatrix}
    \begin{pmatrix}
        \frac{1}{4} & \frac{1}{4} & \frac{1}{4} & \frac{1}{4} & \frac{1}{4} & \frac{1}{4} & \frac{1}{4} & \frac{1}{4} & \frac{1}{4} & \frac{1}{4} & \frac{1}{4}\\
        \frac{1}{4} & \frac{1}{4} & \frac{1}{4} & \frac{1}{4} & \frac{1}{4} & \frac{1}{4} & \frac{1}{4} & \frac{1}{4} & \frac{1}{4} & \frac{1}{4} & \frac{1}{4}\\
        \frac{1}{4} & \frac{1}{4} & \frac{1}{4} & \frac{1}{4} & \frac{1}{4} & \frac{1}{4} & \frac{1}{4} & \frac{1}{4} & \frac{1}{4} & \frac{1}{4} & \frac{1}{4}\\
        \frac{1}{4} & \frac{1}{4} & \frac{1}{4} & \frac{1}{4} & \frac{1}{4} & \frac{1}{4} & \frac{1}{4} & \frac{1}{4} & \frac{1}{4} & \frac{1}{4} & \frac{1}{4}\\
    \end{pmatrix}
    \end{split}
\end{equation}

\subsubsection{Zonotopes Approximation in Real Space}
The Integer space may be a special scenario. More generally we have to expand into Real space. Let's first look at the linear transformation from $z^1$ to $z^2$. Suppose in concrete domain we have N different points in arbitrary positions $X = \{x_1, x_2, ..., x_N\}$, and $X$ is over-approximated by $z^1$. If we lift $X$ up into probability space with uniform distribution we have $d(X) = \{\frac{1}{N}, \frac{1}{N}, ..., \frac{1}{N}\}$. In abstract domain, we know in mathematics there are infinitely many points in Real space contained in $z^1$, but we can regard this as a very large number and let it be $\lim_{M \rightarrow \infty}M$. Then, we can have
\begin{equation}
    A \in \lim_{M \rightarrow \infty} \mathbb{R}^{M \times N}
\end{equation}
with an entry 1 in each column in corresponding position and other entries are all zeros. And 
\begin{equation}
    G \in \lim_{M \rightarrow \infty} \mathbb{R}^{N \times M}
\end{equation}
with all entries are $\frac{1}{N}$. Now we can let the same very large number $\lim_{M \rightarrow \infty} M$ to be the number of all real points contained in $z^2$, so
\begin{equation}
    A' = A, G' = G
\end{equation}
As it is linear transformation, again we have $\overrightarrow{f}: \mathcal{V}(\mathbb{R}^2) \rightarrow \mathcal{V}(\mathbb{R}^2)$
\begin{equation}
    \overrightarrow{f}(d) = d
\end{equation}
and hence $f^{\#}: \mathcal{V}(\mathbb{R}^2) \rightarrow \mathcal{V}(\mathbb{R}^2)$
\begin{equation}
    f^{\#} = A'\overrightarrow{f}G = A'G      
\end{equation}
In ReLU transformation, we have to consider $ReLU(f(X))$. We claim that if $\forall x_{i}, x_{j} \in ReLU(f(X))$ s.t. $x_{i1} \neq x_{j1} \wedge x_{i2} \neq x_{j2}$, then $\overrightarrow{ReLU}: \mathcal{V}(\mathbb{R}^2) \rightarrow \mathcal{V}(\mathbb{R}^2)$
\begin{equation}
    \overrightarrow{ReLU}(d) = d
\end{equation}
and
\begin{equation}
    A'' = A', G'' = G'
\end{equation}
Otherwise, suppose if $\exists$ and only $\exists$ a pair $x_{p}, x_{q} \in ReLU(f(X)) (p < q)$ s.t. $x_{p2} = x_{q2}$, then 
\begin{equation}
    \overrightarrow{ReLU} = 
    \begin{pmatrix}
        1 & 0 & . & . & . & . & . & . & . & .\\
        0 & 1 & . & . & . & . & . & . & . & .\\
        . & . & . & . & . & . & . & . & . & .\\
        . & . & . & . & . & . & . & . & . & .\\
        . & . & . & 1 & . & . & 1 & . & . & .\\
        . & . & . & . & . & . & . & . & . & .\\
        . & . & . & . & . & . & . & . & 1 & 0\\
        . & . & . & . & . & . & . & . & 0 & 1\\
    \end{pmatrix}
\in \mathbb{R}^{(N-1) \times N} 
\end{equation}
which can be obtained by eliminating the $q$th row from identity matrix and assign $\overrightarrow{ReLU}_{pq} = 1$. That is
\begin{equation}
    \overrightarrow{ReLU}     
    \begin{pmatrix}
    \frac{1}{N}\\
    \frac{1}{N}\\
    .\\
    .\\
    .\\
    .\\
    .\\
    \frac{1}{N}\\
    \end{pmatrix}
    = 
    \begin{pmatrix}
    \frac{1}{N}\\
    \frac{1}{N}\\
    .\\
    \frac{2}{N}\\
    .\\
    .\\
    \frac{1}{N}\\
    \end{pmatrix}
\end{equation}
with $\frac{2}{N}$ in $p$th entry. Then
\begin{equation}
    A'' \in \lim_{M \rightarrow \infty} \mathbb{R}^{M \times (N-1)}
\end{equation}
with an entry 1 in each column in corresponding position and other entries are all zeros. And 
\begin{equation}
    G'' \in \lim_{M \rightarrow \infty} \mathbb{R}^{(N-1) \times M}
\end{equation}
with all entries are $\frac{1}{N-1}$. Therefore
\begin{equation}
    ReLU^{\#} = A''\overrightarrow{ReLU}G' 
\end{equation}
Obviously if $\exists$ $r$ exclusive pairs $x_{p}, x_{q} \in ReLU(f(X))$ s.t. $x_{p2} = x_{q2}$, then 
\begin{equation}
    \overrightarrow{ReLU} \in \mathbb{R}^{(N-r) \times N}
\end{equation}
and if let $N \rightarrow M$ we have
\begin{equation}
    \lim_{N \rightarrow M}\overrightarrow{ReLU} \in \mathbb{R}^{(\frac{M}{2}) \times M}
\end{equation}

\subsection{Grids Approximation as Abstract Domain}
We see that in Real space we encounter problem of infinity - since there are infinitely many points in any closed area in Real space, the dimensions of both input space and abstract element tends to infinitely large, and hence we are forced to handle vectors and matrices in dimension of infinity. We always wish to avoid this. A possible approximation is to use grids. That is, in former example, instead of considering space $\mathbb{R}^2$, we consider $\epsilon\mathbb{Z}^2$, where $\epsilon$ gives size of grids to be chosen. For example, if we let $\epsilon$ to be 0.01, then the grid size is $0.01 \times 0.01$, which means we will have $100 \times 100$ points in a closed space bounded by a Box of size $1 \times 1$(regardless of points on boundary lines - we could have $101 \times 101$ points more specifically). In this way, each point represents the local probability of neighbour area, which still shows probability density distribution of the space. For convenience we always let $x_i - axis$ to be one of lines of grids, which will determine position of the whole grids. Now if we consider $z^1$ in $\epsilon\mathbb{Z}^2$, we have in total $30301$ points in this abstract element. That is, recall the notation $M$ used in last subsection, instead of having $\lim_{M \rightarrow \infty}M$ points, we now have $M = 30301$. Same as the input space $X$ in $\epsilon\mathbb{Z}^2$. No matter what $X$ is, $N$ is finite and computable. Therefore we have $A: \mathcal{V}(0.01\mathbb{Z}^2) \rightarrow \mathcal{V}(0.01\mathbb{Z}^2)$
\begin{equation}
    A \in \mathbb{R}^{M \times N} = \mathbb{R}^{30301 \times N}
\end{equation}
and so on for $G$. As we know that any linear transformation in $\mathbb{R}^2$ is a bijective mapping, and $\epsilon\mathbb{Z}^2 \subset \mathbb{R}^2$, hence any linear transformation in $\epsilon\mathbb{Z}^2$ is also a bijective mapping. So we can have $\overrightarrow{f}: \mathcal{V}(\epsilon\mathbb{Z}^2) \rightarrow \mathcal{V}(\epsilon\mathbb{Z}^2)$
\begin{equation}
    \overrightarrow{f}(d) = d
\end{equation}
Similarly $z^2$ in $\mathcal{V}(\epsilon\mathbb{Z}^2)$ contains $60401$ points, so
\begin{equation}
    A' \in \mathbb{R}^{M \times N} = \mathbb{R}^{60401 \times N}
\end{equation}
and so on for $G'$. Then we have $f^{\#}: \mathcal{V}(\epsilon\mathbb{Z}^2) \rightarrow \mathcal{V}(\epsilon\mathbb{Z}^2)$
\begin{equation}
    f^{\#} = A'\overrightarrow{f}G = A'G      
\end{equation}
Similar for $ReLU^{\#}$ that we can now replace $\lim_{M \rightarrow \infty}M$ by finite and computable numbers. The reason $AI^2$ takes zonotopes as abstract domain is that it is concerning robustness of a neural network, and having a sound yet precise abstract element which over-approximates input space is a good way to prove robustness. In probabilistic abstract interpretation, we usually have a different aim. Probabilistic abstract interpretation concerns more about density distribution of input space and its change through propagation of a neural network. Therefore, it is more meaningful to use other abstract domains to reach the purpose. Following grid approximation, using a more sparse grids as an abstract domain is a fair choice. In this context let $z^1$ in $0.01\mathbb{Z}^2$ be input space, i.e., $X = \{x \in z^1 \mid x \in 0.01\mathbb{Z}^2)\}$. Then we let the abstract domain to be the space in $\mathbb{Z}^2$. We can choose different shapes of grids. As an instance, we choose the squares. That is, we abstract 10,000 points in each $1 \times 1$ square in $0.01\mathbb{Z}^2$ into one single point in the centre of the square, and the value of probability of which is the sum of all values of probabilities of 10,000 points. Say if we consider the coordinates $x \in (-3, 3)$ and $y \in (-3, 3)$, then
\begin{equation}
    A \in \mathbb{R}^{49 \times 490,000}, G \in \mathbb{R}^{490,000 \times 49}
\end{equation}
and A' and G' are similar. A concrete element is a vector of length 490,000 and an abstract element is a vector of length 49, with each entry being value of probabilities. We can see that an advantage of grid approximation is that it saves computation complexity. $\overrightarrow{f}$ is very computation consuming as 490,000 affine transformation calculations are needed, while $f^{\#}$ only does 49 such calculations. To lift up $f$ to $\overrightarrow{f}$, we do affine transformation $f$ and the result point will be the position of entry of $\overrightarrow{f}$. Besides squares, we can also have different shape of grids such as diamonds. Different shapes lead to different approximations of summation of probabilities.

\section{Experiment}
In this section we give an example which takes MNIST hand-written digit classifier to illustrate analysis of probabilistic abstract interpretation on convolutional neural network. 
\subsection{The Dataset}
We obtain an open-source dataset `MNIST Digit Dataset' from Kaggle https://www.kaggle.com/competitions/digit-recognizer. The dataset contains about 40,000 handwritten images of digits. Each image is 28 pixels in height and 28 pixels in width, for a total of 784 pixels in total. Each pixel has a single pixel-value associated with it, indicating the lightness or darkness of that pixel, with higher numbers meaning darker. This pixel-value is an integer between 0 and 255, inclusive.

\subsection{The Model}
We design our model to be an eight-layer convolutional neural network, comprising of two $28 \times 28 \times 32$ convolutional layers with $5 \times 5$ filter followed by a $2 \times 2$ max-pooling layer, and then another two $14 \times 14 \times 64$ convolutional layers with $3\times 3$ filter followed by a $2 \times 2$ max-pooling layer, and finally two dense layers, respectively. In total the network contains 887,530 parameters. We choose RMSprop as our Optimizer and categorical cross entropy as our Loss Function. In final epoch of training, our model achieved metrics 0.8917 and loss 0.3446. 
\begin{figure}
\begin{subfigure}{0.5\textwidth}
  \centering
  \includegraphics[width=\linewidth]{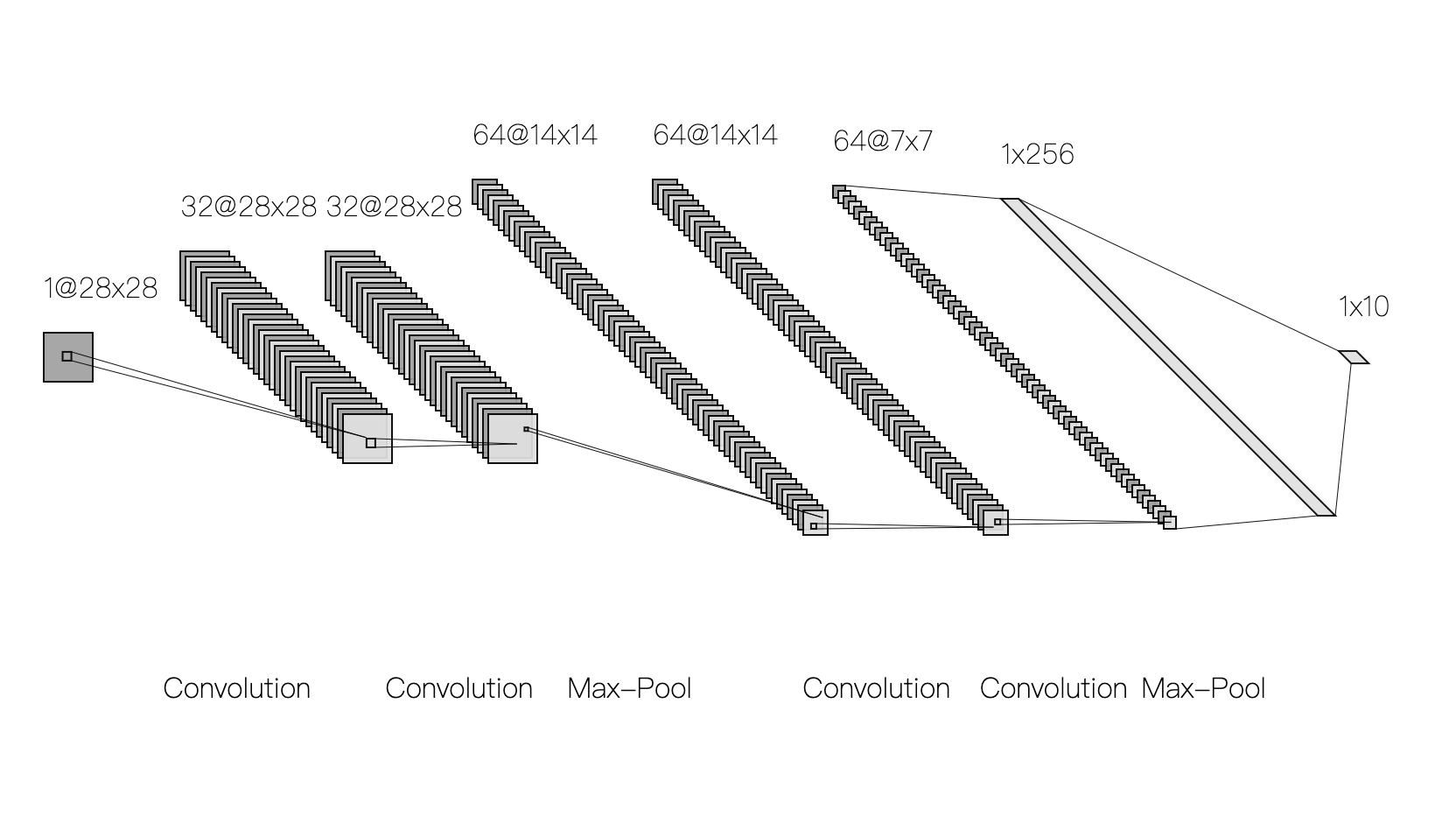}
  \label{fig:sub1}
  \subcaption{Model Architecture Plot}
\end{subfigure}%
\hspace*{\fill}
\begin{subfigure}{0.5\textwidth}
  \centering
  \includegraphics[width=\linewidth]{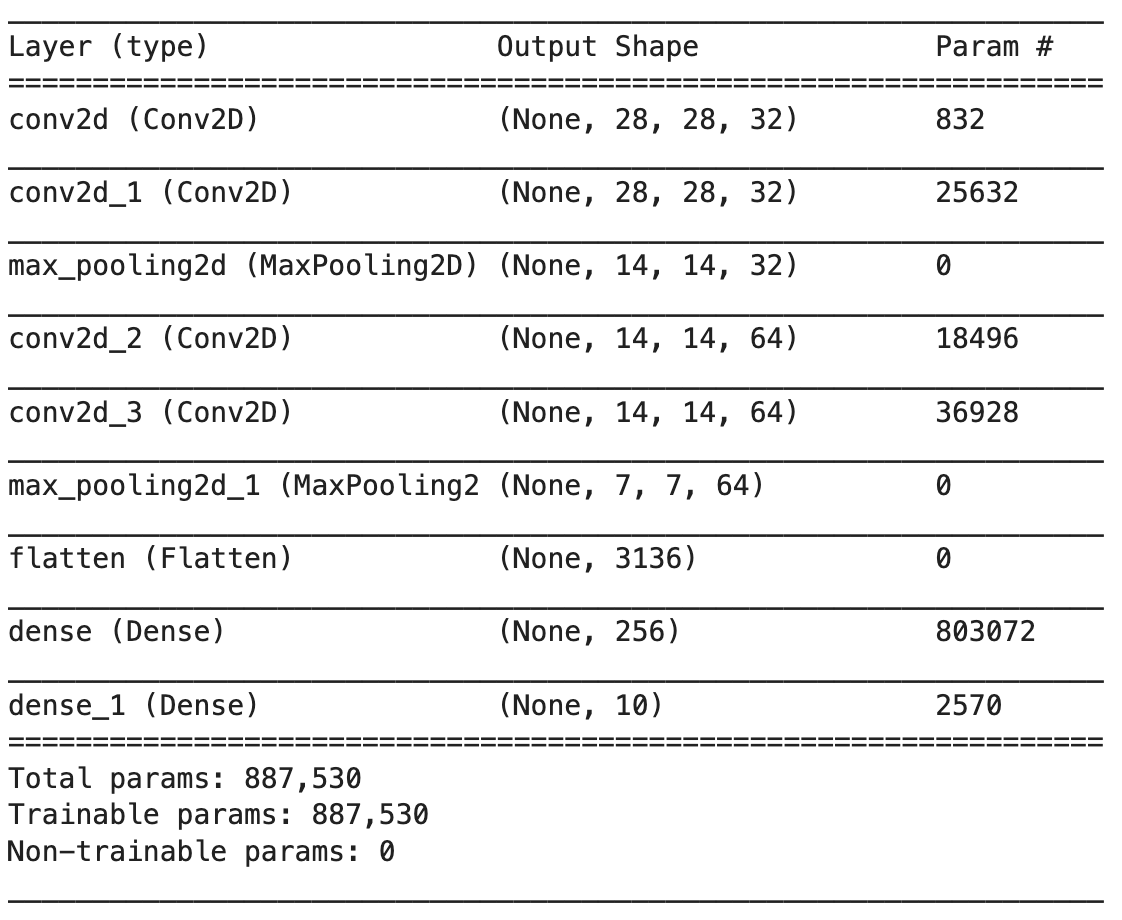}
  \label{fig:sub2}
  \subcaption{Model Architecture Table}
\end{subfigure}
\caption{(a) A plot of model architecture of digit classifier. (b) A table of model architecture showing total number of weights and number of weights in each layer.}
\label{fig:test}
\end{figure}

\subsection{Abstract Domain}
An input image has 784 pixels and each pixel has 256 pixel values. So there are $256^{784}$ possible input images and we cannot exhaustively run them. We take grids approximation as our abstract domain. For each pixel, we classify its pixel-value into simply two categories, being value of 0 as dark and 255 as bright respectively. we also combine a neighborhood of $7 \times 7$ pixels with the same pixel-value, that is, we divide an image into 16 groups of pixels. A grid in this case is 16-dimensional, with each dimension has two choices of values. Therefore in total we have $2^{16}$ grids in this approximation. It shows that for example the network will more likely recognize a bright top-left corner as a feature of digit `3', and will more likely recognize a bright bottom-right corner as a feature of digit `2' or `9'.

\subsection{Distribution Initialisation}
We can sample density initialisation of inputs from training set. In this approximation we have $2^{16}$ grids. We approximate each of training image into corresponding grid and calculate sample probability of each grid with respect to the training set. The prediction of a high probability grid is more likely to occur. 

\subsection{Evaluation}
\cite{ai2} gives an example of classical abstract interpretation analyzer to prove robustness against brightening attack on MNIST digit classifier. The analyzer is able to prove whether a network does correct classification under attack which perturbs an image by changing all pixels above a particular threshold to the brightest possible value. In this case, probabilistic abstract interpretation aims for different purpose. It (i)extracts features of the network and (ii)shows probability density flow of all possible inputs towards prediction. 

\section{Conclusion and Future Work}
We have discussed applying theory of probabilistic abstract interpretation on various kinds of neural network layers. We can see probabilistic abstract interpretation can give information on density distribution flow of possible inputs, which in particular cases is an advantage comparing to classical abstract interpretation analysis framework that is sometimes sound but not precise enough.\\
During the discussion we can also see that problems remained in our approach. The examples we illustrated were mostly in low dimension, but in neural networks it is very common to be in very high dimensions. Secondly, we should discuss more kinds of abstract domains and abstract transformers, in addition to grids approximation, in order to fit different application situations. We remain these as future works.




\end{document}